\documentclass[10pt,twocolumn,letterpaper]{article}

\usepackage[pagenumbers]{cvpr} 

\usepackage{graphicx}
\usepackage{amsmath}
\usepackage{booktabs}

\usepackage{url}
\usepackage{lmodern}
\usepackage{mathtools, amssymb}
\usepackage{float}
\usepackage{xint} 
\usepackage{xinttools} 
\usepackage{tabularx}
\usepackage{multirow}
\usepackage{subcaption}
\usepackage[ruled]{algorithm2e}
\usepackage{newtxtext}
\usepackage[toc,page]{appendix}
\usepackage{makecell}

\usepackage{array}

\SetAlCapHSkip{0pt}
\makeatletter
\newcommand{\thealgorithm}{\arabic\algocf@float}

\newcommand{\AlgoCaptionFormat}{}
\newcommand{\SetAlgoCaptionFormat}[1]{\def\AlgoCaptionFormat{#1}}

\renewcommand{\algocf@makecaption@ruled}[2]{%
  \global\sbox\algocf@capbox{\hskip\AlCapHSkip%
    \setlength{\hsize}{\columnwidth}
    \addtolength{\hsize}{-2\AlCapHSkip}
    \vtop{\AlgoCaptionFormat\algocf@captiontext{#1}{#2}}}
}%

\patchcmd{\@algocf@start}
  {-1.5em}
  {-0.2em}
  {}{}
\robustify\@latex@warning@no@line
\@namedef{ver@everyshi.sty}{}
\makeatother
\usepackage{tikz}
\usetikzlibrary{decorations,intersections,calc,arrows.meta}
\SetAlgoCaptionFormat{\centering\small}

\newcommand{\ubar}{\tikz[overlay] \draw (0,1em)--(0,0em);}
\newcommand{\dbar}{\tikz[overlay] \draw (0,.5em)--(0,-1em);}
\newcommand{\abar}{\tikz[overlay] \draw (0,0.8em)--(0,0em);}
\newcommand{\na}[1]{\makebox[#1]{-}}
\newcommand{\laln}[2]{\makebox[#1]{\hfill#2}}

\newcommand{\mysetminusD}{\hbox{\tikz{\draw[line width=0.6pt,line cap=round] (3pt,0) -- (0,6pt);}}}
\newcommand{\mysetminusT}{\mysetminusD}
\newcommand{\mysetminusS}{\hbox{\tikz{\draw[line width=0.45pt,line cap=round] (2pt,0) -- (0,4pt);}}}
\newcommand{\mysetminusSS}{\hbox{\tikz{\draw[line width=0.4pt,line cap=round] (1.5pt,0) -- (0,3pt);}}}
\newcommand{\mysetminus}{\mathbin{\mathchoice{\mysetminusD}{\mysetminusT}{\mysetminusS}{\mysetminusSS}}}

\usepackage[pagebackref,breaklinks,colorlinks]{hyperref}

\usepackage[capitalize]{cleveref}
\crefname{section}{Sec.}{Secs.}
\Crefname{section}{Section}{Sections}
\Crefname{table}{Table}{Tables}
\crefname{table}{Tab.}{Tabs.}
\Crefname{algorithm}{Algorithm}{Algorithms}
\crefname{algorithm}{Alg.}{Algs.}

\usepackage{authblk}

\def\cvprPaperID{5740} 
\def\confName{CVPR}
\def\confYear{2023}

\begin{document}

\title{Slimmable Pruned Neural Networks}

\author{Hideaki Kuratsu}
\author{Atsuyoshi Nakamura}
\affil{Hokkaido University}
\affil{\tt\small \{hideaki,atsu\}@ist.hokudai.ac.jp}
\maketitle

\begin{abstract}
  Slimmable Neural Networks (S-Net) is a novel network which enabled to select one of the predefined proportions of channels (sub-network) dynamically depending on the current computational resource availability. The accuracy of each sub-network on S-Net, however, is inferior to that of individually trained networks of the same size due to its difficulty of simultaneous optimization on different sub-networks. In this paper, we propose Slimmable Pruned Neural Networks (SP-Net), which has sub-network structures learned by pruning instead of adopting structures with the same proportion of channels in each layer (width\footnote{Width refers to the number of channels in a layer.} multiplier) like S-Net, and we also propose new pruning procedures: \textbf{multi-base pruning} instead of one-shot or iterative pruning to realize high accuracy and huge training time saving. We also introduced \textbf{slimmable channel sorting (scs)} to achieve calculation as fast as S-Net and \textbf{zero padding match (zpm) pruning} to prune residual structure in more efficient way. SP-Net can be combined with any kind of channel pruning methods and does not require any complicated processing or time-consuming architecture search like NAS models. Compared with each sub-network of the same FLOPs on S-Net, SP-Net improves accuracy by 1.2-1.5\% for ResNet-50, 0.9-4.4\% for VGGNet, 1.3-2.7\% for MobileNetV1, 1.4-3.1\% for MobileNetV2 on ImageNet. Furthermore, our methods outperform other SOTA pruning methods and are on par with various NAS models according to our experimental results on ImageNet. The code is available at \textsl{\url{https://github.com/hideakikuratsu/SP-Net}}.
\end{abstract}

\section{Introduction}
  Deep Neural Network (DNN) is one of the promising machine learning methods which has been achieving state-of-the-art in a variety of fields. However, the model size of it has been becoming larger, which has been also increasing its prediction cost with its higher accuracy. In contrast to it, IoT products or compact devices began to spread rapidly in recent years, and they need more and more responsive and accurate AI systems such as self-driving car, real-time translation, or surveillance camera, thus a lot of researchers have been working on these demands.

  \textit{Pruning} \cite{han:prune,li:convnets,liu:slimming,liu:metapruning,he:softfilter,he:chprune,luo:thinet} is one of the most powerful compressing methods which can dramatically reduce computational cost and the amount of parameters compared to other methods. However, pruning itself is used in advance to get slim models which meet constraints of small devices, and it does not take real-time conditions of the devices into account. For instance, IoT devices may go into power save mode, or run other power-intensive processes. Thus, we need DNN that can adjust computational power or memory usage in response to the device states \textit{dynamically}.

  Adaptive Computation Graph \cite{teerapittayanon:branchynet,liu:d2nn,yu:slimmable,yu:universally} are the methods that has such dynamic adaptivity. Especially, Yu \etal~\cite{yu:slimmable,yu:universally} proposed \textit{Slimmable Neural Networks} (\textit{S-Net}), which is an architectural design that can select one of various sized sub-networks with shared parameters depending on current computational/memory loads. Although existing Adaptive Computation Graph methods made it possible to change network capacity dynamically, instead many of them failed to achieve accuracy on par with pruning methods.

  In recent years, there are some research that work on the performance issues using weight-sharing NAS \cite{cai:onceforall,pham:paramsharing} inspired by above methods. These works enabled networks to achieve high accuracy while learning in much shorter time than existing NAS, but still it is much slower compared to methods without NAS.

  Therefore, we worked on bridging between those works in order to realize high accuracy, dynamic adaptivity, and shorter training time, and propose improved S-Net based on our new methods, named \textit{Slimmable Pruned Neural Networks} (\textit{SP-Net}). In this paper, we propose \textit{multi-base pruning}, \textit{slimmable channel sorting} (\textit{scs}), and \textit{zero padding match (zpm) pruning} as core methods of SP-Net.

  As one of the core methods of SP-Net, we propose \textit{multi-base pruning}: we first train some individual base networks for pruning, then embed the pruned architectures (number of channels to use in each layer) of each base network into \(\times\)1.0 base network after pruning. This method is different from other single-base methods like one-shot or iterative pruning in that our method uses multiple base networks for pruning. With our method, we can parallelize the pruning process by training base networks individually instead of repeating pruning and fine-tuning process in consecutive way. Additionally, multi-base pruning prunes and fine-tunes just once, thus it saves time-consuming fine-tuning on SP-Net.

  However, it still has some problems for practical use without additional methods, since SP-Net usually has non-uniformly pruned structures as sub-networks. First, non-uniform pruning on SP-Net causes random accesses on the memory since non-contiguous channels are usually selected in each layer by pruning, but non-selected channels cannot be removed on SP-Net because all of the channels in each layer are used by largest network. This leads to devastatingly slow computation, thus we proposed \textit{slimmable channel sorting} (\textit{scs}) to avoid the random accesses by sorting channels in each layer of \(\times\)1.0 base network according to pruning precedences (\cref{fig:channelsorting}). With this method, we realized to minimize the latency increasing within around 3ms compared to S-Net of the same size (\cref{ablation:resnet50:latency}). Second, pruning ResNet-like structure \cite{he:resnet,sandler:mobilenetv2} is a bit tricky with non-uniform pruning because it requires index match of filters at the point where shortcut path joins. Existing papers \cite{liu:metapruning,li:convnets,luo:thinet,liu:slimming} worked on this issue, but they failed to propose an effective way. We introduced more efficient way of pruning when we prune the meeting point, which we call \textit{zero padding match (zpm) pruning}: supplementing filter index mismatch with zero-filled channels on pruning, and realized significant accuracy improvement (0.5-2.4\% for ResNet-50 on ImageNet) compared to model without it.

  By combining all of the core methods of SP-Net, we achieved remarkable improvement to S-Net. On ImageNet dataset, SP-Net improved in accuracy by 1.2-1.5\% for ResNet-50, 0.9-4.4\% for VGGNet, 1.3-2.7\% for MobileNetV1, 1.4-3.1\% for MobileNetV2 compared to each sub-network of the same FLOPs on S-Net. On CIFAR-100, SP-Net realized 2.8-3.3\% improvement for ResNet-50 and 0.8-3.7\% for VGGNet. Furthermore, we compared SP-Net and other pruning methods. SP-Net consistently outperformed other pruning methods including MetaPruning \cite{liu:metapruning} which is one of the SOTA pruning methods by 1.9\% for MobileNetV1 and 1.1\% for MobileNetV2 on ImageNet at lower FLOPs. Additionally, we also compared SP-Net and various NAS models. SP-Net did not outperform all of them, but achieved accuracy on par with them without any search cost. As an ablation study, we compared results between with or without the core methods of SP-Net.

  \begin{figure}[t]
    \centering
    \includegraphics[width=1.0\linewidth,clip]{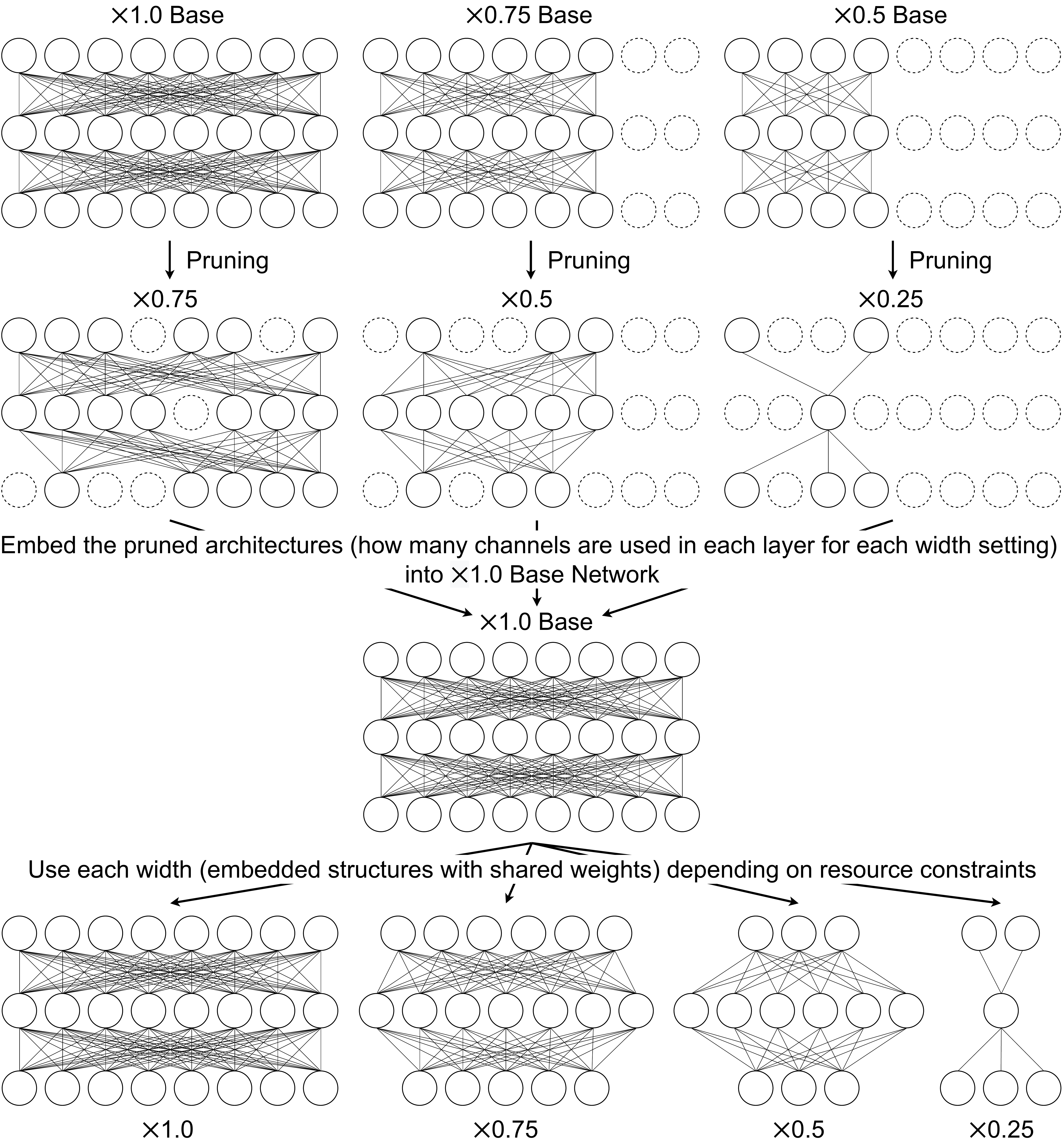}
    \vspace{-1.8em}
    \caption{Illustration of constituting SP-Net with multi-base pruning. We first train \#[width settings]\(-\)1 base networks for pruning, then prune these networks and embed the pruned architectures (how many channels are used in each layer) into \(\times\)1.0 base network. After fine-tuning or retraining from scratch in the same way as S-Net, it can be used for selecting appropriate sub-networks and computing dynamically according to applications or device states.}
    \label{fig:multibasepruning}
  \end{figure}

  \section{Related Work}
  \noindent
  \textbf{Weight Pruning.} There are various methods to compress DNN such as Low-rank Decomposition \cite{denton:lowrank}, Quantization \cite{han:deepcompress,rastegari:xnornet,hubara:binarized}, and Distillation \cite{hinton:distill,xie:noisystudent,wang:alphanet}. Among them, weight pruning is the very effective method which zero or remove less important weights under specific criteria. Magnitude-based unstructured pruning \cite{han:prune,han:deepcompress} prunes the weights with smaller \(L_1\) value. Although it's very simple and effective, it makes sparse parameters, and it is generally difficult to gain computational acceleration with sparse representation without dedicated libraries or hardware. Thus channel pruning is the secondly effective and promising method. It is not so efficient as magnitude-based pruning since it prunes coarser than magnitude-based one, but it requires no specific libraries or hardware to gain computational acceleration, because it does not make sparse parameters by removing each channel entirely. Li \etal~\cite{li:convnets} introduced \(L_1\)-norm based channel pruning which prunes channels with smaller sum of \(L_1\) value in each layer. Luo \etal~\cite{luo:thinet} minimized reconstruction errors using statistics of layer \(i + 1\) to prune layer \(i\) in a greedy manner. These methods prune networks based on manually predefined pruning ratios in each layer, thus it's not trivial that how many channels you should prune to achieve high compression rate, and it can also cause devastating performance degradation by pruning a layer overly. Liu \etal~\cite{liu:slimming} introduced global channel pruning, called \textit{network slimming} which automates the decision of pruning ratios in each layer using commonly used Batch Normalization \cite{ioffe:bn}, and achieved SOTA compression performance on some models and datasets. We used network slimming as a pruning method of SP-Net in our experiments due to its effectiveness and simpleness.

  \noindent
  \textbf{Adaptive Computation Graph.} It is used to reduce computational costs by changing network architectures dynamically. Teerapittayanon \etal~\cite{teerapittayanon:branchynet} used branch structures that calculate entropy to determine when to terminate the computation. Graves \cite{graves:act} introduced \textit{halting score} which determines when to stop computation based on cumulative sum of it. Liu and Deng \etal~\cite{liu:d2nn} introduced controller modules to select appropriate route to calculate outputs dynamically based on the difficulty of input data. Yu \etal~\cite{yu:slimmable,yu:universally} proposed a novel architecture, called \textit{Slimmable Neural Networks} (\textit{S-Net}), which enables to compute adaptively by selecting a sub-network of appropriate computational costs depending on running applications or device states. However, inference accuracies of S-Net are relatively low compared to individual models of the same size. Thus we incorporated non-uniform pruning methods to compose architectures of sub-networks on S-Net with our methods of SP-Net to boost the performance.

  \noindent
  \textbf{Neural Architecture Search (NAS).} NAS automates finding good architectures which yields high performance on specified tasks within limited search space using algorithms like evolutionary algorithm \cite{liu:hierarchical,liu:metapruning,real:amoebanet}, reinforcement learning, \cite{zoph:naswithreinforcementlearning,tan:mnasnet,zoph:nasnet}, Bayesian Optimization \cite{kandasamy:baysianoptimisation}. Its searching costs are usually very high (10k-2000k GPU hours in general) with its vast search space. Pham \etal~\cite{pham:paramsharing} proposed weight-sharing method by representing search space as directed acyclic graph (DAG) and found architectures as subgraphs of DAG. In recent years, some research are also incorporating this weight-sharing idea in order to limit search space and save costs for searching \cite{wang:attentivenas,yu:bignas,cai:onceforall}, but instead weight-sharing NAS takes lots of training costs because it repeats pruning and fine-tuning process many times. Not only this process cannot be parallelized, but also requires time-consuming simultaneous fine-tuning of largest network and its sub-networks which shares weights with it. SP-Net also uses weight-sharing training, but requires pruning and fine-tuning process just once, and with multi-base pruning, we can parallelize the pruning process by training multiple base networks. Thus SP-Net takes at most \(\simeq 2\times\) as much as S-Net in training time without any search cost while achieving high performance on par with those NAS models. For feature comparison between SP-Net and other methods, refer to \cref{ablation:analysis}.

  \begin{table*}
    \footnotesize
    \centering
    \begin{tabular}{ccccc} \hline
      Method & Accuracy & Search cost & Training cost & Adaptivity \\ \hline
      pruning & Middle to High & Low or Zero & Low to High & \\
      S-Net & Low & Zero & Low & \checkmark \\
      NAS & High & High & Low to Middle & \\
      weight-sharing NAS & High & Low to Middle & High & \(\ast\) \\
      SP-Net & High & Zero & Low & \checkmark \\ \hline
    \end{tabular}
    \vspace{-0.5em}
    \caption{Comparison of each method which yields slim models. \(\ast\)~represents not all of the models have adaptivity.}
    \label{ablation:analysis}
    \vspace{-1.5em}
  \end{table*}

  \vspace{-0.5em}
  \section{Slimmable Pruned Neural Networks}
  First, we explain Slimmable Neural Networks (S-Net), then describe the core methods of Slimmable Pruned Neural Networks (SP-Net), then finally show the whole algorithm of training SP-Net.

  \subsection{Slimmable Neural Networks}
  \label{snet}
  Slimmable Neural Networks (S-Net) \cite{yu:slimmable} is a novel network that enables to use multiple sub-networks as different computational routes depending on the resource constraints by changing the number of channels to use in each layer. We first define width settings list like \(\times\)[0.25, 0.5, 0.75, 1.0] as what proportion of channels are used in each layer (width multiplier), \eg under \(\times\)0.25 setting, we use only 25\% channels of each layer to calculate feature maps. For implementation details, refer to \cref{appnd:snet}. S-Net embeds multiple sub-networks into one network, thus we can switch these sub-networks to calculate outputs depending on the state of the device, and save the amount of parameters by weight-sharing without preparing multiple networks of each size. The training algorithm of S-Net is described in \cref{alg:1}\footnote{They also use \textit{Switchable Batch Normalization} in order to converge the validation loss of S-Net. For detailed information, refer to \cite{yu:slimmable}.}.

  \begin{algorithm}[t]
    \small
    \SetAlgoNoEnd
    \SetAlgoNoLine
    \NoCaptionOfAlgo
    \SetKwProg{Until}{until}{ do}{end}
    \SetKwFor{RepTimes}{repeat}{iterations}{end}
    \SetKw{KwForIn}{in}
    \textbf{Parameters:} \\
    \Indp
    \hspace{-0.7em}\(W = [w_1, \ldots, w_k]\): width (proportion of channels) list \\
    \Indm
    \RepTimes{\(n\)}{
      \hspace{-0.7em}Get batch of data \(x\) and label \(y\)\;
      \hspace{-0.7em}\For{\(w\) \KwForIn \(W\)}{
        \hspace{-1.3em}\(M_w\): sub-network of \(M\) with \(\times w\) channels in each layer\;
        \hspace{-1.3em}Calculate outputs: \(\hat{y} = M_w(x)\)\;
        \hspace{-1.3em}Compute loss from (\(y,\, \hat{y}\))\;
        \hspace{-1.3em}Compute gradients from loss\;
      }
      \hspace{-0.7em}Update the parameters of \(M\) from stacked gradients\;
    }
    \caption{\textbf{Algorithm 1} Training S-Net \(M\)}
    \label{alg:1}
  \end{algorithm}

  \subsection{Multi-Base Pruning}
\begin{figure}
    \centering
    \includegraphics[width=0.97\linewidth,clip]{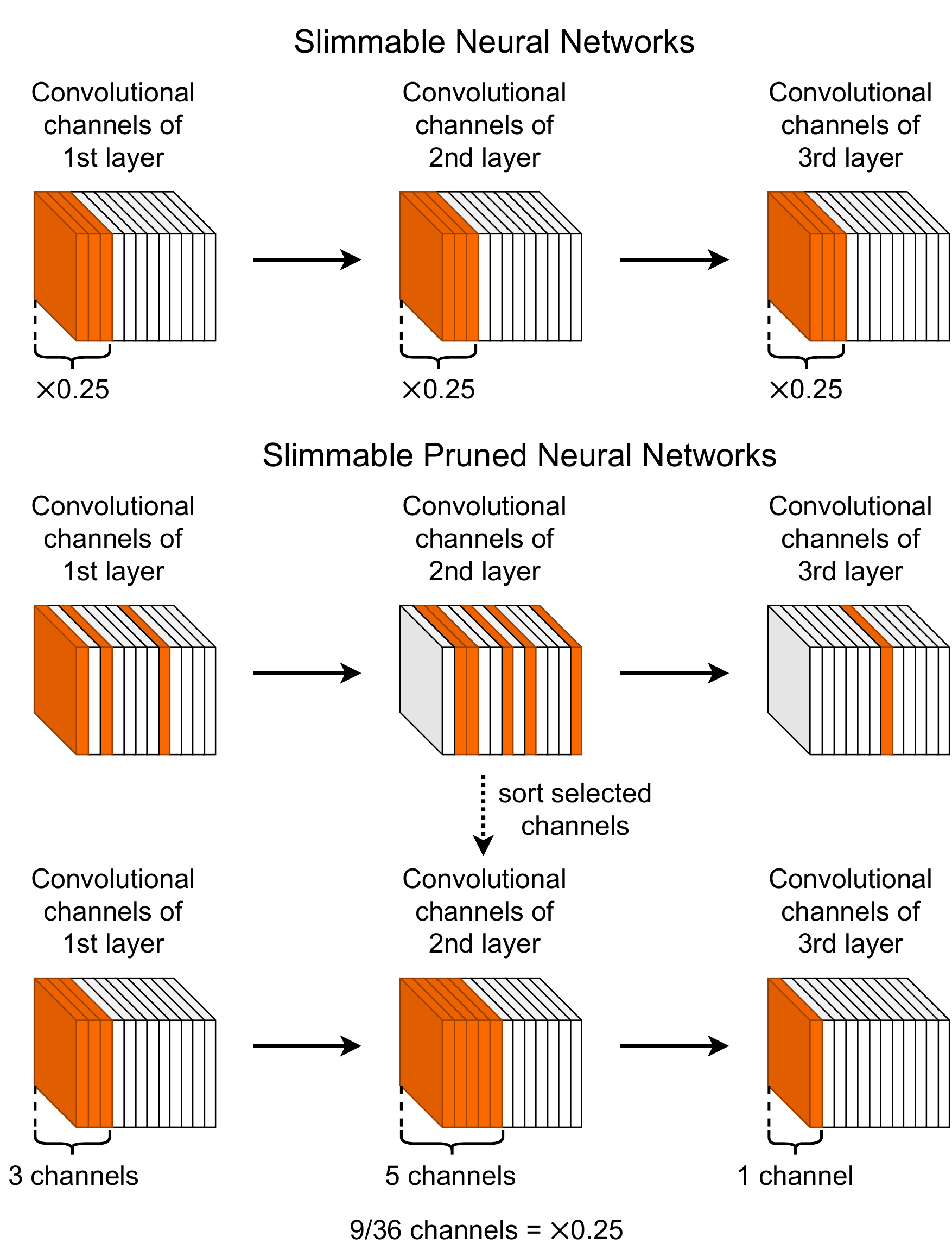}
    \vspace{-0.5em}
    \caption{Illustration of difference of how to select channels in each layer between S-Net and SP-Net. Colored channels represent selected ones in the layer. This figure shows \(\times\)0.25 width setting.}
    \label{fig:pruningdifference}
  \end{figure}

  In \cref{snet}, we described that S-Net selects channels of each layer uniformly. This method is very simple and easy to implement, but it does not take relative importance of each channels into consideration, thus it usually leads to lower accuracy compared to other methods which select more important channels according to its pruning criteria. Hence we incorporated pruning methods into S-Net for selecting sub-network architectures. We show the difference between S-Net and SP-Net when we select channels of a sub-network under \(\times\)0.25 width setting in \cref{fig:pruningdifference}. Note that width settings on SP-Net represent the proportions of channels in \textit{all} of the layers, not in \textit{per} layer due to global pruning.

  In previous methods \cite{li:convnets,liu:slimming,yu:autoslim,luo:thinet,he:chprune}, pruning has been done in single-base way, that is, after training one large network from scratch, the network is pruned using one-shot or iterative \cite{liu:slimming,han:prune,frankle:lottery} pruning according to pruning precedences. One-shot pruning literally determines which channels to choose on all of the layers at a time. Thus it takes no time cost, but it is generally inferior in accuracy to iterative way. On the other hand, iterative pruning repeats the channel-decision process various times to select channels in more refined way. Thus it takes huge amount of time, since it usually needs fine-tuning at least three times to achieve high accuracy \cite{frankle:lottery}, and it cannot be parallelized due to its consecutive way of pruning. Additionally, fine-tuning or retraining on SP-Net after pruning takes much more time than training individual networks of the same FLOPs, because non-uniformly pruned structures are less efficient in memory access than uniform ones. Therefore, we propose more well-balanced way of pruning that resolve both of the performance issues and slow training, which we call \textit{multi-base pruning}. In this method, we first train \#[width settings]\(-\)1 base networks by selecting the same proportion of channels from each layer (\eg train \(\times 0.5, \times 0.75, \times 1.0\) base networks individually for \(\times\)[0.25, 0.5, 0.75, 1.0] setting). Then we prune these base networks to obtain pruned architectures of the same FLOPs as each width setting (\eg prune \(\times 0.5\) base network and obtain pruned architectures of the same FLOPs as \(\times 0.25\) network). Finally, we embed these architectures into \(\times\)1.0 base network as its sub-networks (as shown in the bottom of \cref{fig:pruningdifference}, we store only \#channels information, \eg under \(\times\)0.25 setting, we use three channels for layer 1 of \(\times\)1.0 base network, five channels for layer 2, \textellipsis). It is illustrated in \cref{fig:multibasepruning}. With multi-base pruning, we can save vast amount of training time since training of multiple base networks can be parallelized, and time-consuming fine-tuning or retraining on SP-Net is done just once. Additionally, multi-base pruning improved accuracy of SP-Net dramatically in our experiments (Sec. \hyperlink{ablation:multibase}{\ref*{ablation}}).

  \subsection{Slimmable Channel Sorting (SCS)}
\begin{figure}
    \centering
    \includegraphics[width=1.0\linewidth,clip]{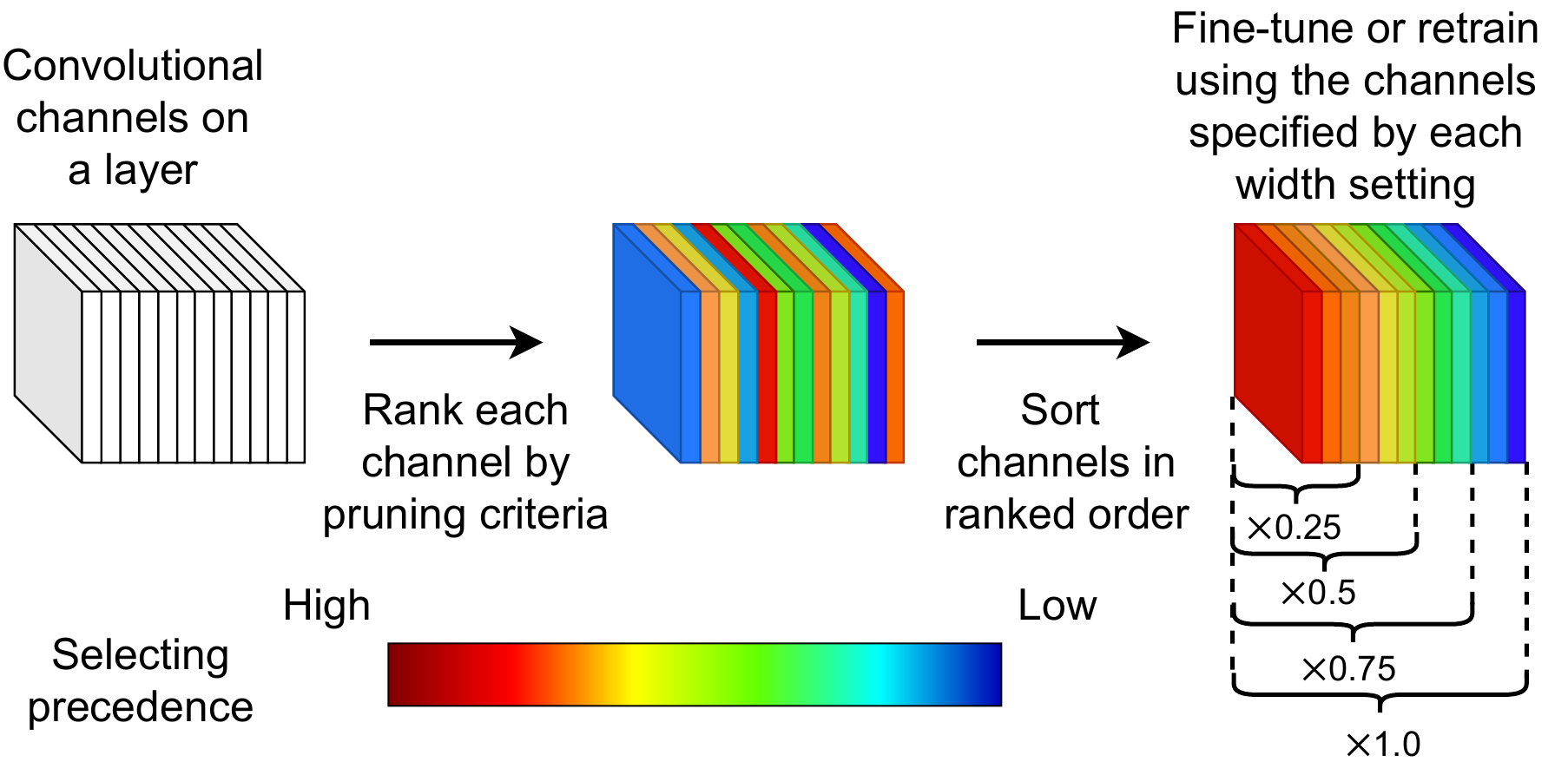}
    \vspace{-1.8em}
    \caption{Illustration of scs on a layer. This figure shows sorting of filters on a layer (\(i\)), but it actually sorts input channels of next layer (\(i + 1\)) as well in order not to change the outputs of whole network. Note that contiguous channels are always used like S-Net.}
    \label{fig:channelsorting}
  \end{figure}

  In our experiments, we found sub-networks on SP-Net which have pruned architectures are very slow in computation (2\(\times\) in worst case) compared to S-Net of the same size. This is because S-Net always uses adjacent channels (thus, very efficient due to its sequential access on memory) since it uniformly selects channels in each layer, while SP-Net does not necessarily use adjacent ones since it selects according to its pruning criteria (middle of \cref{fig:pruningdifference}). On SP-Net, non-selected channels cannot be removed like individually pruned networks since the parameters of sub-networks are always shared with larger ones. Thus, inefficient random memory access occurs on SP-Net. Therefore, we proposed \textit{slimmable channel sorting} (\textit{scs}): sort channels in each layer of \(\times\)1.0 base network according to the pruning precedence to force the sub-networks to always use adjacent channels even after non-uniform pruning. It's illustrated in the bottom of \cref{fig:pruningdifference} and \cref{fig:channelsorting}.

  Channel sorting itself is already used in other papers \cite{li:convnets,cai:onceforall}, but scs is different from them in that it not only sorts filters in layer (\(i\)), but also input channels in layer (\(i + 1\)) to avoid changes of outputs of network, in other words, channel sorting has never been used for speeding up computation by avoiding random memory access while keeping the same outputs. With this method, we achieved almost the same calculation speed as S-Net (within 3ms in our experiments). For concrete comparison results, refer to Sec. \hyperlink{ablation:sorting}{\ref*{ablation}}.

  \subsection{Zero Padding Match (ZPM) Pruning}
  \label{sec:zpm}
\begin{figure}[t]
    \centering
    \subfloat[][pruning uniformly]{\includegraphics[width=0.8\linewidth,clip]{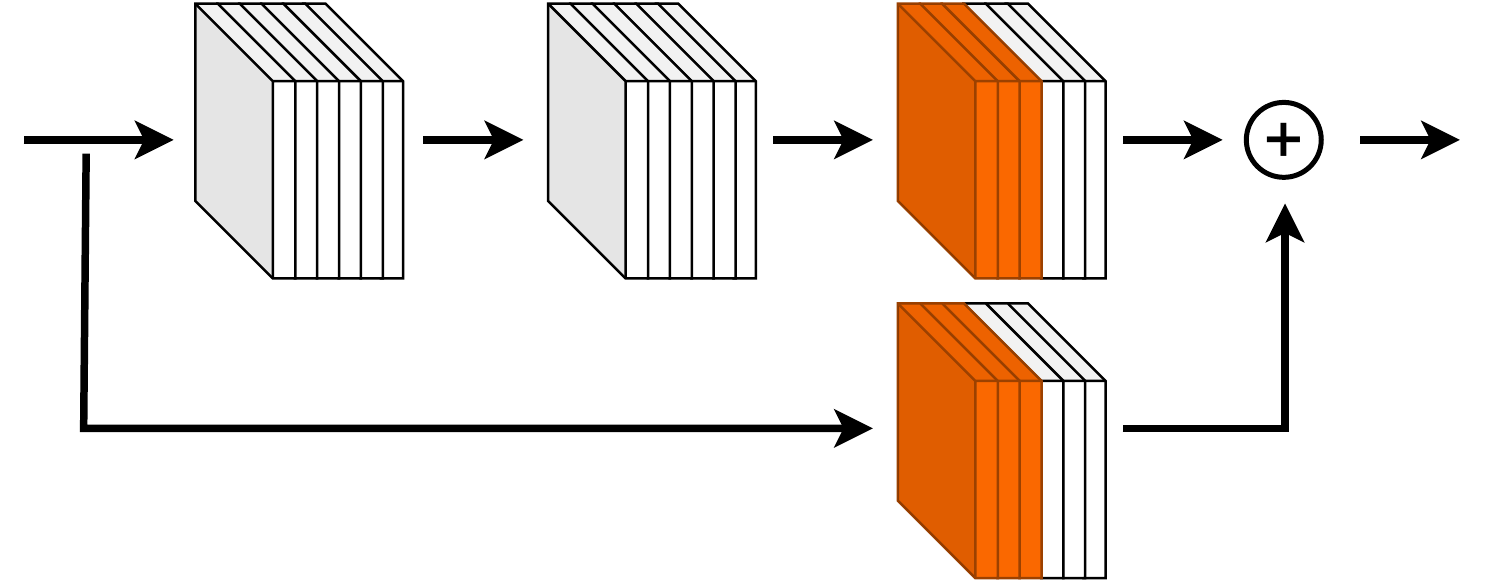}\label{fig:pruningshortcut:1}}

    \subfloat[][prioritize shortcut path]{\includegraphics[width=1.0\linewidth,clip]{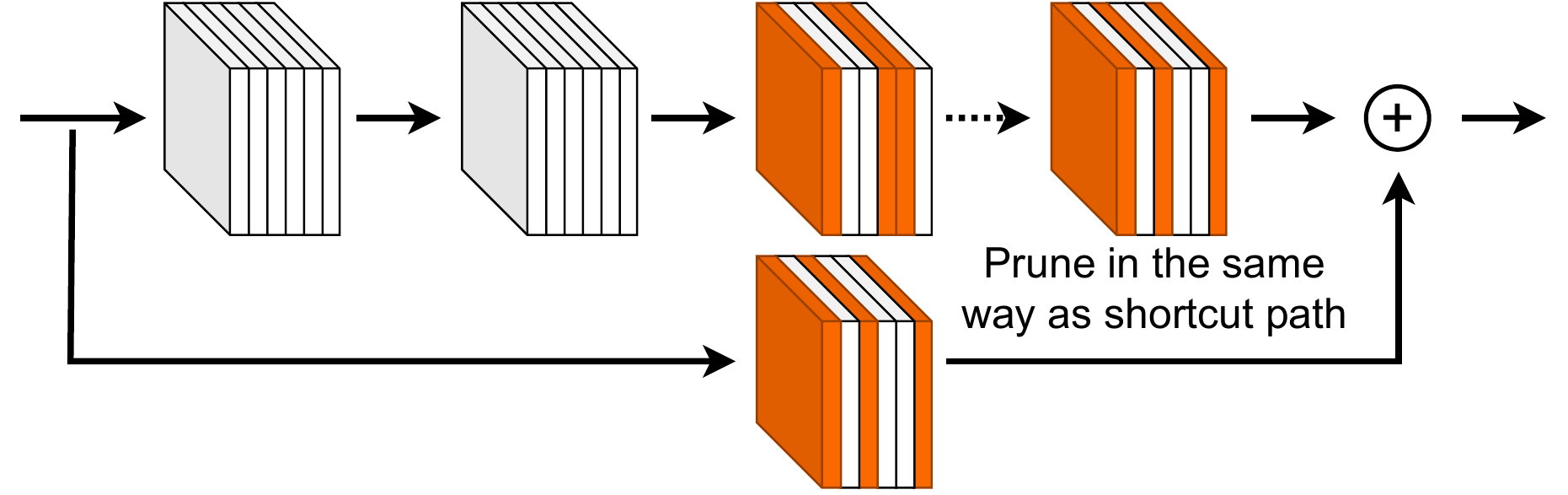}\label{fig:pruningshortcut:2}}

    \subfloat[][zpm pruning]{\includegraphics[width=1.0\linewidth,clip]{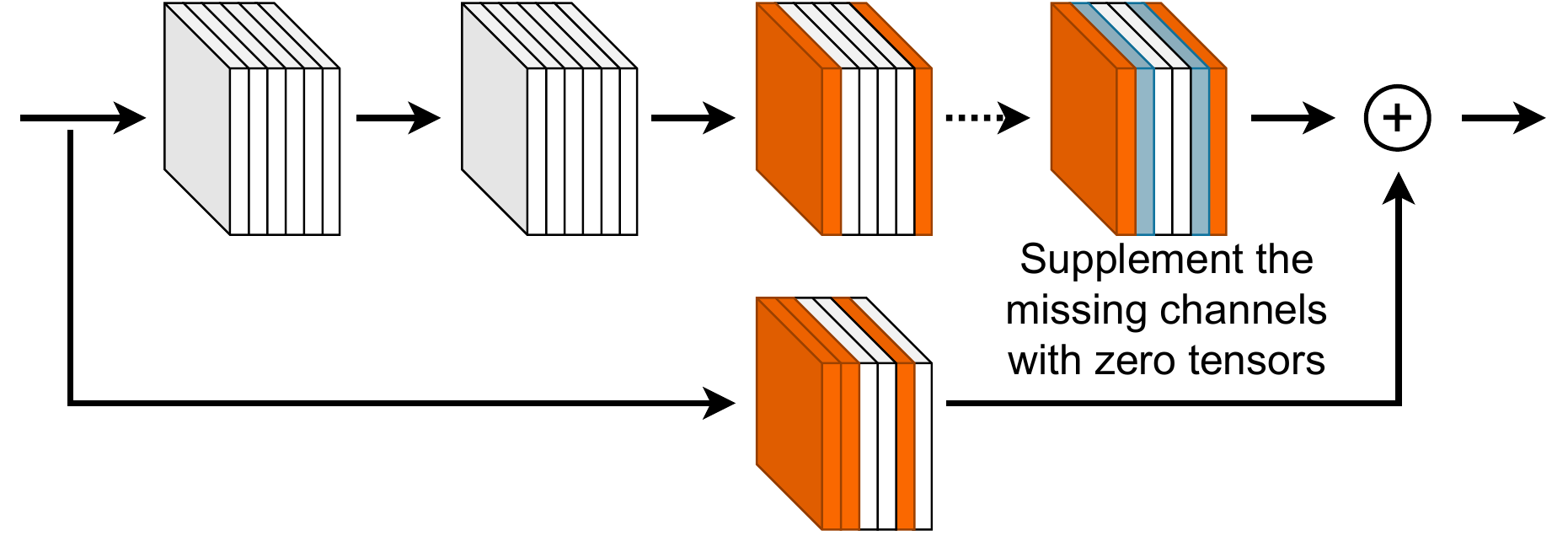}\label{fig:pruningshortcut:3}}
    \vspace{-0.5em}
    \caption{Difference of how to prune at confluence where shortcut path joins on residual block. Selected channels are represented by orange color, and supplemented channels with zero tensors are represented by blue color. \cref{fig:pruningshortcut:1}: prune both paths uniformly and identically. \cref{fig:pruningshortcut:2}: prioritize the way of pruning on shortcut path and overwrite the way of pruning on non-shortcut path with it. \cref{fig:pruningshortcut:3}: keep the way of pruning on both paths and fill missing channels with zeros. Solid arrow means a unit of processes (convolution, Batch Normalization, and activation) from layer \(L\) to \(L + 1\), and dashed arrow means changes of pruning pattern in the layer \(L\). Note that the dashed arrows do not mean a new layer is added, but mean penultimate layer is pruned in last one's way.}
    \label{fig:pruningshortcut}
  \end{figure}

  ResNet \cite{he:resnet} is one of the most famous networks, and well known for its residual structure (or skip connection). This structure solved the vanishing/exploding gradients issues and enabled deeper architectural design and high performance, so it's now widely used on a variety of networks \cite{radosavovic:regnet,tan:efficientnet,sandler:mobilenetv2,tan:mnasnet,zhang:shufflenet}. For ResNet, there's a point to note for pruning the place where shortcut path joins: indices of channels on two path have to be matched since add operations of outputs from these path occurs. There's been mainly three ways for this pruning: (1) Not pruning at all \cite{luo:thinet}, (2) pruning both path uniformly and identically \cite{yu:slimmable,yu:universally,liu:metapruning} and (3) prioritize pruning pattern of shortcut path \cite{li:convnets,liu:slimming}. For (1), we do not prune the confluence at all, thus we can't prune so many channels, and it also leads to lower accuracy because lots of less important channels remain. For (2) (\cref{fig:pruningshortcut:1}), it's simple and effective if the network is trained from scratch because we do not have to care about the importance (remaining precedence) of each channel when we do not fine-tune. In fact, this method is used in \cite{yu:slimmable,yu:universally,liu:metapruning} since they train networks from scratch, but this method can't be used for fine-tuning, and also causes inefficiency because it forces both path to be pruned in the same way. For (3) (\cref{fig:pruningshortcut:2}), we always prioritize the way shortcut path is pruned and overwrite the pruning pattern of the non-shortcut path with it. This method was first suggested by Li \etal~\cite{li:convnets} based on the idea that shortcut path conveys more important information than the other path, and in fact achieved high performance on ResNet, then other research also used the same method \cite{liu:slimming}. However, we found that although this method works well on relatively easier datasets like CIFAR-10 or CIFAR-100 \cite{dataset:cifar}, it does not on more difficult datasets like ImageNet \cite{dataset:imagenet}, because it isn't fully optimized due to overwriting of pruning pattern of non-shortcut path. Thus we propose more efficient way of pruning the residual structure, \textit{zero padding match (zpm) pruning}: we keep the pruning pattern on both paths while filling missing channels with zeros, then add up outputs from these paths as normal (\cref{fig:pruningshortcut:3}). This method requires additional masks (only when we also use scs) and temporary zero tensors for add operations (Refer to \cref{appnd:zpm} for implementation), and FLOPs increases compared to other methods, but we realized significant accuracy improvement with it even if we consider FLOPs increasing. In ablation study, we compare method (3) and zpm pruning in Sec. \hyperlink{ablation:zpm}{\ref*{ablation}}.

  \subsection{Slimmable Pruned Neural Networks}
  Finally, we show the whole training algorithm of SP-Net in \cref{alg:2}. For implementation details including pruning, refer to \cref{appnd:spnet,appnd:pruning}.

  \begin{algorithm}[t]
    \small
    \SetAlgoNoEnd
    \SetAlgoNoLine
    \NoCaptionOfAlgo
    \SetKwProg{Until}{until}{ do}{end}
    \SetKwFor{RepTimes}{repeat}{iterations}{end}
    \SetKw{KwForIn}{in}
    \textbf{Parameters:} \\
    \Indp
    \hspace{-0.7em}\(W = [w_1, \ldots, w_k]\): width (proportion of channels) list \\
    \Indm
    \For{\(w_i\) \KwForIn \(W \mysetminus \left\{w_1\right\}\)}{
      \hspace{-0.7em}Train individual model \(M_{w_i}\) with \(\times w_i\) channels in each layer\;
      \hspace{-0.7em}Prune \(M_{w_i}\) and obtain pruned architectures \(\mathcal{A}_{w_{i-1}}\) of the same FLOPs as \(M_{w_{i-1}}\)\;
    }
    Embed \(\{\mathcal{A}_{w_1}, \ldots, \mathcal{A}_{w_{k-1}}\}\) into \(M_{w_k}\) as sub-networks\;
    Sort channels in each layer of \(M_{w_k}\) with pruning precedence\;
    \RepTimes{\(n\)}{
      \hspace{-0.7em}Get batch of data \(x\) and label \(y\)\;
      \hspace{-0.7em}\For{\(w\) \KwForIn \(W\)}{
        \hspace{-1.3em}\(M_w\): sub-network of \(M_{w_k}\) corresponding to \(w\)\;
        \hspace{-1.3em}Calculate outputs: \(\hat{y} = M_w(x)\)\;
        \hspace{-1.3em}Compute loss from (\(y,\, \hat{y}\))\;
        \hspace{-1.3em}Compute gradients from loss\;
      }
      \hspace{-0.7em}Update the parameters of \(M_{w_k}\) from stacked gradients\;
    }
    \caption{\textbf{Algorithm 2} Training SP-Net \(M\) (\(= M_{w_k}\))}
    \label{alg:2}
  \end{algorithm}

  \vspace{-0.5em}
  \section{Experiments}
  In this section, we first show the comparison results between individually trained networks, S-Net, and SP-Net, then we show the ablation study using core methods of SP-Net. For heavy models, we used ResNet-50\cite{he:resnet} and VGGNet\cite{simonyan:vgg} on CIFAR-100\cite{dataset:cifar} and ImageNet \cite{dataset:imagenet}. For mobile models, we used MobileNetV1 \cite{howard:mobilenets} and MobileNetV2 \cite{sandler:mobilenetv2} on ImageNet. We used \textit{network slimming} \cite{liu:slimming} as a pruning method. For both of datasets, we used 224\(\times\)224 size images and followed official implementation of ImageNet training examples of Pytorch \cite{code:imagenet} for augmentation settings. For CIFAR-100 results, we reported average of five runs.

  \vspace{-0.3em}
  \subsection{Performance comparison}
  \label{sec:performance}
  \noindent
  \textbf{Training details for heavy models on CIFAR-100.} For training of individual models, S-Net, and base networks of SP-Net, we used batch size 128 and trained 120 epochs with initial learning rate 0.1, and multiplied by 0.2 at 60/90/110 epoch, and used SGD with weight decay \(10^{-4}\), Nesterov momentum 0.9 without dampening, and sparsity rate \(\lambda = 10^{-4}\) for VGGNet and \(\lambda = 10^{-5}\) for ResNet-50. For SP-Net of ResNet-50, we used \(\times0.75, \times1.0\) base networks to train with \(\times\)[0.5, 0.75, 1.0] settings, and fine-tuned 60 epochs with initial learning rate 0.1, and multiplied by 0.1 at 50 epoch. We used \(T = 1\) and \(\alpha = 0.9\) for inplace knowledge distillation. The rest of the settings are the same as training from scratch. For SP-Net of VGGNet, we used \(\times0.5, \times0.75\) base networks to train with \(\times\)[0.25, 0.5, 1.0] settings, and the rest of the settings are the same as ResNet-50.

  \vspace{0.3em}
  \noindent
  \textbf{Training details for heavy models on ImageNet.} For training of individual models, S-Net, and base networks of SP-Net, we used batch size 256 and trained 90 epochs with initial learning rate 0.1, and multiplied by 0.2 at 30/50/70 epoch, and used SGD with weight decay \(10^{-4}\), Nesterov momentum 0.9 without dampening, and sparsity rate \(\lambda = 10^{-5}\) for ResNet-50 and VGGNet. For training of SP-Net, we fine-tuned for ResNet-50 and retrained from scratch for VGGNet, because VGGNet achieved higher performance when retrained than fine-tuned. The reason of this difference in accuracy has been studied in many research \cite{liu:rethink,frankle:lottery,cai:onceforall}, but we would like to work on this in the future. In our experiments, we found that models with residual structure achieve higher accuracy when fine-tuned, and ones without it achieve higher when retrained for both of heavy and mobile models. We used \(\times0.75, \times1.0\) base networks for ResNet-50 to train SP-Net with \(\times\)[0.5, 0.75, 1.0] settings and \(\times0.5, \times0.75\) base networks for VGGNet to train SP-Net with \(\times\)[0.25, 0.5, 1.0] settings same as on CIFAR-100, and fine-tuned 60 epochs with initial learning rate 0.01, and multiplied by 0.1 at 30/45 epoch for both models. For inplace knowledge distillation, we used \(T = 1\) and \(\alpha = 0.9\). The rest of the settings are the same as training from scratch.

  \vspace{0.3em}
  \noindent
  \textbf{Training details for mobile models on ImageNet.} For training of individual models, S-Net, and base networks of SP-Net, we followed Li's settings \cite{li:mobilenetv2imp} to train MobileNetV1 and MobileNetV2. For training of SP-Net, we fine-tuned MobileNetV2 and retrained MobileNetV1 150 epochs with initial learning rate 0.08 without warm-up. For MobileNetV1, we used \(\times0.75, \times1.0, \times1.3\) base networks to train with \(\times\)[0.5, 0.75, 1.0, 1.5] settings, and \(T = 2\) and \(\alpha = 0.9\) for inplace knowledge distillation. For MobileNetV2, we used \(\times1.0, \times1.3\) base networks to train with \(\times\)[0.75, 1.0, 1.3] settings, and \(T = 1\) and \(\alpha = 0.9\) for inplace knowledge distillation.

  \vspace{0.3em}
  \noindent
  \textbf{Results for heavy models.}
  We show the results of heavy models (ResNet-50 and VGGNet) on CIFAR-100 in \cref{result:cifar}. SP-Net outperformed individually/slimmable trained networks by noticeable margin at every FLOPs (2.7-3.0\% for ResNet-50 and 0.2-3.6\% for VGGNet compared to individual models). For ImageNet (\cref{result:imagenet}), SP-Net is inferior to the individual models under \(\times\)1.0 for VGGNet, but still outperformed S-Net significantly and even individual models at all of the rest FLOPs (1.6-4.1\%), and ResNet-50 with our methods outperformed them at every FLOPs (0.3-2\%). For both of models, our methods improved accuracy significantly especially at lower FLOPs. Furthermore, we compared the results of ResNet-50 on ImageNet using our methods and AutoSlim \cite{yu:autoslim} which is one of the NAS methods. As shown in \cref{result:comparenas}, our methods realized high performance on par with AutoSlim. Note that all of these performance enhancements were achieved only by channel selection using a pruning method without time-consuming architecture search.

  \vspace{0.3em}
  \noindent
  \textbf{Results for mobile models.}
  We show the results of mobile models (MobileNetV1 and MobileNetV2) on ImageNet in \cref{result:imagenet}. SP-Net outperformed individually/slimmable trained networks by noticeable margin at every FLOPs (0.8-2.9\% for MobileNetV1 and 0.7-2.6\% for MobileNetV2). We also compared SP-Net and various NAS models in \cref{result:comparenas}. Our models such as SP-MobileNetV2 did not outperform all of these NAS models, but achieved accuracy on par with them only by a pruning method which takes no search cost as shown in the last column of \cref{result:comparenas}. In these experiments, we used fixed pruning method, base models, and so on. But our methods also have flexibility that can be combined with any kind of channel pruning methods \cite{liu:metapruning,li:convnets,he:chprune,he:softfilter,luo:thinet} and any type of promising models \cite{zhang:shufflenet,tan:mnasnet,tan:efficientnet,radosavovic:regnet,cai:onceforall} as well as recently proposed self-attention mechanism for S-Net \cite{li:dsnet} or new knowledge distillation methods for weight-sharing models \cite{wang:alphanet}.

  \begin{table}
    \footnotesize
    \centering
    \begin{tabular}{ccrrc} \Xhline{2\arrayrulewidth}
      Model & Method & Params & FLOPs & Error (\%) \\ \Xhline{2\arrayrulewidth}
      \multirow{9}{*}{ResNet-50} & \hspace{-0.5em}\laln{2em}{\(\times\)1.0}  & 23.7M & 4.1G & 24.5 \\
                                 & \hspace{-0.5em}\laln{3em}{\(\times\)0.75} & 13.4M & 2.3G & 24.6 \\
                                 & \hspace{-0.5em}\laln{2em}{\(\times\)0.5}  &  6.0M & 1.1G & 25.3 \\ \cline{2-5}
                                 & \multirow{3}{*}{S}                        & 23.8M & 4.1G & 24.6 \\
                                 &                                           & 13.4M & 2.3G & 24.9 \\
                                 &                                           &  6.0M & 1.1G & 25.6 \\ \cline{2-5}
                                 & \multirow{3}{*}{\shortstack{SP \\(ours)}} & 23.7M & 4.1G & \textbf{21.8} \\
                                 &                                           & 21.1M & 2.3G & \textbf{21.9} \\
                                 &                                           & 11.1M & 1.1G & \textbf{22.3} \\ \Xhline{2\arrayrulewidth}
      \multirow{9}{*}{VGGNet}    & \hspace{-0.5em}\laln{2em}{\(\times\)1.0}  & 20.1M & 19.6G & 22.9 \\
                                 & \hspace{-0.5em}\laln{2em}{\(\times\)0.5}  &  5.0M &  4.9G & 24.4 \\
                                 & \hspace{-0.5em}\laln{3em}{\(\times\)0.25} &  1.3M &  1.2G & 27.5 \\ \cline{2-5}
                                 & \multirow{3}{*}{S}                        & 20.1M & 19.6G & 23.5 \\
                                 &                                           &  5.0M &  4.9G & 24.5 \\
                                 &                                           &  1.3M &  1.2G & 27.6 \\ \cline{2-5}
                                 & \multirow{3}{*}{\shortstack{SP \\(ours)}} & 20.1M & 19.6G & \textbf{22.7} \\
                                 &                                           &  9.5M &  4.9G & \textbf{22.9} \\
                                 &                                           &  3.3M &  1.2G & \textbf{23.9} \\ \Xhline{2\arrayrulewidth}
    \end{tabular}
    \vspace{-0.9em}
    \caption{Top-1 test errors of heavy models on CIFAR-100. Each scaling factor in Method represents width multiplier of channels as in \cite{howard:mobilenets,sandler:mobilenetv2}, and S and SP indicates S-Net and SP-Net. Classification errors are reported from best test errors during training.}
    \label{result:cifar}
  \end{table}

  \begin{table}
    \footnotesize
    \centering
    \begin{tabular}{ccrrc} \Xhline{2\arrayrulewidth}
      Model & Method & Params & FLOPs & Error (\%) \\ \Xhline{2\arrayrulewidth}
      \multirow{9}{*}{ResNet-50}                     & \hspace{-0.5em}\laln{2em}{\(\times\)1.0}  & 25.6M &  4.1G & 23.7 \\
                                                     & \hspace{-0.5em}\laln{3em}{\(\times\)0.75} & 14.8M &  2.3G & 24.9 \\
                                                     & \hspace{-0.5em}\laln{2em}{\(\times\)0.5}  &  6.9M &  1.1G & 27.8 \\ \cline{2-5}
                                                     & \multirow{3}{*}{S}                        & 25.6M &  4.1G & 24.6 \\
                                                     &                                           & 14.8M &  2.3G & 25.3 \\
                                                     &                                           &  6.9M &  1.1G & 27.3 \\ \cline{2-5}
                                                     & \multirow{3}{*}{\shortstack{SP \\(ours)}} & 25.6M &  4.1G & \textbf{23.4} \\
                                                     &                                           & 17.6M &  2.3G & \textbf{23.9} \\
                                                     &                                           & 10.3M &  1.1G & \textbf{25.8} \\ \Xhline{2\arrayrulewidth}
      \multirow{9}{*}{VGGNet}                        & \hspace{-0.5em}\laln{2em}{\(\times\)1.0}  & 20.5M & 19.6G & \textbf{25.0} \\
                                                     & \hspace{-0.5em}\laln{2em}{\(\times\)0.5}  &  5.3M &  4.9G & 30.9 \\
                                                     & \hspace{-0.5em}\laln{3em}{\(\times\)0.25} &  1.4M &  1.2G & 41.1 \\ \cline{2-5}
                                                     & \multirow{3}{*}{S}                        & 20.6M & 19.6G & 27.0 \\
                                                     &                                           &  5.3M &  4.9G & 31.5 \\
                                                     &                                           &  1.4M &  1.2G & 41.4 \\ \cline{2-5}
                                                     & \multirow{3}{*}{\shortstack{SP \\(ours)}} & 20.5M & 19.6G & 26.1 \\
                                                     &                                           &  6.6M &  4.9G & \textbf{29.3} \\
                                                     &                                           &  1.8M &  1.2G & \textbf{37.0} \\ \Xhline{2\arrayrulewidth}
      \multirow{9}{*}{MobileNetV1}                   & \hspace{-0.5em}\laln{2em}{\(\times\)1.0}  &  4.2M &  569M & 28.1 \\
                                                     & \hspace{-0.5em}\laln{3em}{\(\times\)0.75} &  2.6M &  325M & 30.3 \\
                                                     & \hspace{-0.5em}\laln{2em}{\(\times\)0.5}  &  1.3M &  149M & 34.9 \\ \cline{2-5}
                                                     & \multirow{3}{*}{S}                        &  4.3M &  584M & 28.6 \\
                                                     &                                           &  2.6M &  337M & 30.5 \\
                                                     &                                           &  1.3M &  157M & 34.7 \\ \cline{2-5}
                                                     & \multirow{3}{*}{\shortstack{SP \\(ours)}} &  4.9M &  569M & \textbf{27.3} \\
                                                     &                                           &  3.2M &  325M & \textbf{29.0} \\
                                                     &                                           &  1.8M &  150M & \textbf{32.0} \\ \Xhline{2\arrayrulewidth}
      \multirow{9}{*}{MobileNetV2}                   & \hspace{-0.5em}\laln{2em}{\(\times\)1.3}  &  5.4M &  509M & 25.9 \\
                                                     & \hspace{-0.5em}\laln{2em}{\(\times\)1.0}  &  3.5M &  301M & 28.0 \\
                                                     & \hspace{-0.5em}\laln{3em}{\(\times\)0.75} &  2.6M &  209M & 30.3 \\ \cline{2-5}
                                                     & \multirow{3}{*}{S}                        &  5.4M &  509M & 26.6 \\
                                                     &                                           &  3.5M &  301M & 28.5 \\
                                                     &                                           &  2.6M &  209M & 30.8 \\ \cline{2-5}
                                                     & \multirow{3}{*}{\shortstack{SP \\(ours)}} &  5.4M &  509M & \textbf{25.2} \\
                                                     &                                           &  4.0M &  305M & \textbf{26.5} \\
                                                     &                                           &  2.9M &  207M & \textbf{27.7} \\ \Xhline{2\arrayrulewidth}
    \end{tabular}
    \vspace{-0.9em}
    \caption{Top-1 test errors of heavy/mobile models on ImageNet. Each scaling factor in Method represents width multiplier of channels as in \cite{howard:mobilenets,sandler:mobilenetv2}, and S and SP indicates S-Net and SP-Net. Classification errors are reported from best test errors during training.}
    \label{result:imagenet}
    \vspace{-0.4em}
  \end{table}

  \vspace{-0.2em}
  \subsection{Ablation Study}
  \label{ablation}
  In this section, we show the ablation study using ResNet-50 and core methods of SP-Net. For all of the experiments, we used the same training settings as described in \cref{sec:performance}.

  \vspace{0.3em}
  \noindent
  \textbf{Comparison of pruning algorithms.} We compared SP-Net and other pruning methods such as \textit{network slimming} \cite{liu:slimming}, \textit{\(L_1\)-norm based pruning} \cite{li:convnets}, \textit{MetaPruning} \cite{liu:metapruning}, \textit{ThiNet} \cite{luo:thinet}, \textit{CP} \cite{he:chprune}, and \textit{SFP} \cite{he:softfilter} (\cref{ablation:resnet50:pruning}). For \(L_1\)-norm based pruning, we modified it a bit so that it prunes networks globally, which automates decision of pruning ratios in each layer and usually leads to higher accuracy than non-global one \cite{liu:rethink,frankle:lottery}. SP-Net outperformed other pruning methods remarkably especially at lower FLOPs including MetaPruning, which is one of the SOTA pruning methods (by 1.4\% for MobileNetV1 and 1.1\% for MobileNetV2 on ImageNet). In \cref{appnd:pruningcomparison}, we also compared network slimming and \(L_1\)-norm based pruning as pruning methods of SP-Net.

  \vspace{0.3em}
  \noindent
  \hypertarget{ablation:multibase}{\textbf{Comparison using multi-base pruning.}} We compared the results between one-shot, iterative, and multi-base pruning (\cref{ablation:resnet50:multibase}). For iterative pruning, we pruned the \(\times\)1.0 base network so that its FLOPs after pruning is the same as individual model, like \(\times\)1.0 \(\rightarrow \) \(\times\)0.75 \(\rightarrow\) \(\times\)0.5 \(\rightarrow\) \(\times\)0.25, then finally retrained the whole network from scratch as S-Net using the pruned architectures of each size. As shown in the table, multi-base pruning realized significant improvement in accuracy especially at lower FLOPs.

  \vspace{0.3em}
  \noindent
  \hypertarget{ablation:sorting}{\textbf{Comparison using scs.}} We compared inference latencies on NVIDIA V100 GPU with or without scs (\cref{ablation:resnet50:latency}). We can see that scs outstandingly enhances inference speed compared to the model without it; the delay is at most 3.4ms with scs compared to S-Net of the same size while 14.2ms without it, and improved 10.2ms in average. Note that scs never changes outputs of SP-Net.

  \vspace{0.3em}
  \noindent
  \hypertarget{ablation:zpm}{\textbf{Comparison using zpm pruning for ResNet.}} We compared results between with or without zpm pruning on ResNet (\cref{ablation:resnet50:zpm}). The model without zpm pruning uses prioritize-shortcut-path pruning (refer to \cref{sec:zpm}). The results using both zpm pruning and multi-base pruning are also shown in the bottom of the table. We can see zpm pruning noticeably raise accuracies at every FLOPs. Additionally, multi-base pruning further boosts accuracies with zpm pruning (by 4.6\% at lowest FLOPs compared to the model without both of the methods).

  \begin{table}[H]
    \footnotesize
    \centering
    \begin{tabular}{>{\centering\arraybackslash}m{5.8em} >{\centering\arraybackslash}m{5.2em} >{\hspace*{0pt}\hfill}m{2.6em} >{\hspace*{0pt}\hfill}m{2.4em} >{\centering\arraybackslash}m{1.3em} >{\centering\arraybackslash}m{3.2em}} \Xhline{2\arrayrulewidth}
      \shortstack{Model} & Method & Params & FLOPs & \shortstack{Error \\ (\%)} & \vspace{0.2em}\shortstack{Search \\ cost \\ (hours)} \\ \Xhline{2\arrayrulewidth}
      \multirow{4}{*}{ResNet-50}   & \multirow{2}{*}{AutoSlim \cite{yu:autoslim}}  & 20.6M & 2.0G & 24.4 &   900 \\
                                   &                                               & 13.3M & 1.0G & 26.0 & \abar \\ \cline{2-6}
                                   & \multirow{2}{*}{SP (ours)}                    & 17.6M & 2.0G & 24.3 & \textbf{0} \\
                                   &                                               & 10.4M & 1.0G & 26.0 & \abar \\ \Xhline{2\arrayrulewidth}
      \multirow{6}{*}{MobileNetV1} & \multirow{3}{*}{AutoSlim \cite{yu:autoslim}}  &  4.6M & 572M & 27.0 &   180 \\
                                   &                                               &  4.0M & 325M & 28.5 & \dbar \\
                                   &                                               &  1.9M & 150M & 32.1 & \ubar \\ \cline{2-6}
                                   & \multirow{3}{*}{SP (ours)}                    &  4.9M & 569M & 27.3 & \textbf{0} \\
                                   &                                               &  3.2M & 325M & 29.0 & \dbar \\
                                   &                                               &  1.8M & 150M & 32.0 & \ubar \\ \Xhline{2\arrayrulewidth}
      \multicolumn{2}{c}{\multirow{3}{*}{AutoSlim-MobileNetV2 \cite{yu:autoslim}}} &  6.5M & 505M & 24.6 &   180 \\
      \multicolumn{2}{c}{}                                                         &  5.7M & 305M & 25.8 & \dbar \\
      \multicolumn{2}{c}{}                                                         &  4.1M & 207M & 27.0 & \ubar \\ \cline{1-6}
      \multicolumn{2}{c}{\multirow{2}{*}{NasNet \cite{zoph:nasnet}}}               &  5.3M & 564M & 26.0 &   48k \\
      \multicolumn{2}{c}{}                                                         &  5.3M & 488M & 27.2 & \abar \\ \cline{1-6}
      \multicolumn{2}{c}{MNasNet \cite{tan:mnasnet,yu:autoslim}}                   &  4.3M & 317M & 26.0 &   40k \\ \cline{1-6}
      \multicolumn{2}{c}{AmoebaNet \cite{real:amoebanet,liu:darts}}                &  5.1M & 555M & 25.5 &   76k \\ \cline{1-6}
      \multicolumn{2}{c}{\multirow{3}{*}{RegNet \cite{radosavovic:regnet}}}        &  6.1M & 600M & 24.5 &     - \\
      \multicolumn{2}{c}{}                                                         &  4.3M & 400M & 25.9 &     - \\
      \multicolumn{2}{c}{}                                                         &  3.2M & 200M & 29.6 &     - \\ \cline{1-6}
      \multicolumn{2}{c}{ProxylessNAS \cite{cai:proxylessnas,cai:onceforall}}      & \na{2em} & 320M & 25.4 &   200 \\ \cline{1-6}
      \multicolumn{2}{c}{DARTS \cite{liu:darts}}                                   &  4.7M & 574M & 26.7 &    96 \\ \cline{1-6}
      \multicolumn{2}{c}{\multirow{3}{*}{SP-MobileNetV2 (ours)}}                   &  5.4M & 509M & 25.2 & \textbf{0} \\
      \multicolumn{2}{c}{}                                                         &  4.0M & 305M & 26.5 & \dbar \\
      \multicolumn{2}{c}{}                                                         &  2.9M & 207M & 27.7 & \ubar \\ \Xhline{2\arrayrulewidth}
    \end{tabular}
    \vspace{-0.5em}
    \caption{Performance and search cost comparison between various NAS models and SP-Net with top-1 test errors on ImageNet. SP indicates SP-Net. Classification errors are reported from best test errors during training. Other results are cited from referred papers.}
    \label{result:comparenas}
    \vspace{-1.5em}
  \end{table}

  \begin{table}[H]
    \footnotesize
    \centering
    \begin{tabular}{>{\centering\arraybackslash}m{5.8em} >{\centering\arraybackslash}m{7em} >{\hspace*{0pt}\hfill}m{3.6em} >{\hspace*{0pt}\hfill}m{3.4em} >{\centering\arraybackslash}m{2.3em}} \Xhline{2\arrayrulewidth}
      Model & Pruning Method & Params & FLOPs & \vspace{0.2em}\shortstack{Error \\ (\%)} \\ \Xhline{2\arrayrulewidth}
      \multirow{16}{*}{ResNet-50}  & \multirow{3}{*}{\shortstack{SP (ours)}}
                                     & 25.6M & 4.1G & \textbf{23.3} \\
                                   & & 16.0M & 2.0G & \textbf{24.3} \\
                                   & & 10.2M & 1.0G & \textbf{26.0} \\ \cline{2-5}
                                   & \multirow{3}{*}{\shortstack{network \\ slimming \cite{liu:slimming}}}
                                     & 25.6M & 4.1G & 23.6 \\
                                   & & 13.6M & 2.3G & 25.5 \\
                                   & &  6.8M & 1.1G & 31.8 \\ \cline{2-5}
                                   & \multirow{3}{*}{\shortstack{\(L_1\)-norm based \\ pruning \cite{li:convnets}}}
                                     & 25.6M & 4.1G & 23.4 \\
                                   & & 21.2M & 2.3G & 25.2 \\
                                   & & 10.6M & 1.1G & 31.5 \\ \cline{2-5}
                                   & \multirow{2}{*}{MetaPruning \cite{liu:metapruning}}
                                     & \na{3em} & 2.0G & 24.6 \\
                                   & & \na{3em} & 1.0G & 26.6 \\ \cline{2-5}
                                   & \multirow{3}{*}{ThiNet \cite{luo:thinet,liu:metapruning}}
                                     & \na{3em} & 2.9G & 24.2 \\
                                   & & \na{3em} & 2.1G & 25.3 \\
                                   & & \na{3em} & 1.2G & 27.9 \\ \cline{2-5}
                                   & \vspace{0.2em}\shortstack{CP \cite{he:chprune,liu:metapruning}}
                                     & \na{3em} & 2.0G & 26.7 \\ \cline{2-5}
                                   & \vspace{0.2em}\shortstack{SFP \cite{he:softfilter,liu:metapruning}}
                                     & \na{3em} & 2.9G & 24.9 \\ \Xhline{2\arrayrulewidth}
      \multirow{4}{*}{MobileNetV1} & \multirow{2}{*}{MetaPruning \cite{liu:metapruning}}
                                     & \na{3em} & 324M & 29.1 \\
                                   & & \na{3em} & 149M & 33.9 \\ \cline{2-5}
                                   & \multirow{2}{*}{SP (ours)}
                                     &  3.2M & 325M & \textbf{29.0} \\
                                   & &  1.8M & 150M & \textbf{32.0} \\ \Xhline{2\arrayrulewidth}
      \multirow{4}{*}{MobileNetV2} & \multirow{2}{*}{MetaPruning \cite{liu:metapruning}}
                                     & \na{3em} & 313M & 27.3 \\
                                   & & \na{3em} & 217M & 28.8 \\ \cline{2-5}
                                   & \multirow{2}{*}{SP (ours)}
                                     &  4.0M & 305M & \textbf{26.5} \\
                                   & &  2.9M & 207M & \textbf{27.7} \\ \Xhline{2\arrayrulewidth}
    \end{tabular}
    \vspace{-0.5em}
    \caption{Performance comparison between various pruning methods and SP-Net with top-1 test errors on ImageNet. Results of network slimming \cite{liu:slimming} and \(L_1\)-norm based pruning \cite{li:convnets} are reproduced and reported from best test errors during training. Other results are cited from referred papers.}
    \label{ablation:resnet50:pruning}
    \vspace{-1.5em}
  \end{table}

  \begin{table}
    \footnotesize
    \centering
    \begin{tabular}{crrc} \hline
    Network & Params & FLOPs & Error (\%) \\ \hline
    \multirow{4}{*}{\shortstack{SP-ResNet-50 \\(w/ one-shot pruning)}}   & 25.6M & 4.1G & 24.2 \\
                                                                         & 17.6M & 2.3G & 24.8 \\
                                                                         &  9.4M & 1.1G & 27.4 \\
                                                                         &  3.8M & 286M & 38.2 \\ \hline
    \multirow{4}{*}{\shortstack{SP-ResNet-50 \\(w/ iterative pruning)}}  & 25.6M & 4.1G & 24.9 \\
                                                                         & 17.4M & 2.3G & 25.3 \\
                                                                         &  9.6M & 1.1G & 27.8 \\
                                                                         &  3.6M & 286M & 37.4 \\ \hline
    \multirow{4}{*}{\shortstack{SP-ResNet-50 \\(w/ multi-base pruning)}} & 25.6M & 4.1G & \textbf{23.8} \\
                                                                         & 17.6M & 2.3G & \textbf{24.6} \\
                                                                         & 10.3M & 1.1G & \textbf{26.7} \\
                                                                         &  4.0M & 286M & \textbf{35.0} \\ \hline
    \end{tabular}
    \vspace{-0.5em}
    \caption{Performance comparison between three ways of pruning with top-1 test errors of ResNet-50 on ImageNet. Prefix SP indicates SP-Net. Classification errors are reported from best test errors during training.}
    \label{ablation:resnet50:multibase}
    \vspace{-0.5em}
  \end{table}

  \begin{table}
    \footnotesize
    \centering
    \begin{tabular}{crrcc} \hline
      Network & Params & FLOPs & Error (\%) & Latency \\ \hline
      \multirow{4}{*}{S-ResNet-50}                                       & 25.6M & 4.1G & 24.3 & 53.0ms \\
                                                                         & 14.8M & 2.3G & 25.4 & 35.0ms \\
                                                                         &  6.9M & 1.1G & 28.1 & 20.1ms \\
                                                                         &  2.0M & 286M & 35.8 & 10.3ms \\ \hline
      \multirow{4}{*}{\shortstack{SP-ResNet-50 \\(w/o scs)}} & \na{2.8em} & \na{2.1em} & - & 52.7ms \\
                                                                         & \na{2.8em} & \na{2.1em} & - & 45.9ms \\
                                                                         & \na{2.8em} & \na{2.1em} & - & 34.3ms \\
                                                                         & \na{2.8em} & \na{2.1em} & - & 24.3ms \\ \hline
      \multirow{4}{*}{\shortstack{SP-ResNet-50 \\(w/ scs)}}  & 25.6M & 4.1G & 23.8 & 52.7ms \\
                                                                         & 17.6M & 2.3G & 24.6 & 36.9ms \\
                                                                         & 10.3M & 1.1G & 26.7 & 23.5ms \\
                                                                         &  4.0M & 286M & 35.0 & 13.5ms \\ \hline
    \end{tabular}
    \vspace{-0.5em}
    \caption{Hardware latencies of inference by ResNet-50 on NVIDIA V100 GPU with batch size 64. Prefix S indicates S-Net, and SP indicates SP-Net. We omitted some results for SP-ResNet-50 without scs, because they are the same as ones with scs. Classification errors are reported from best test errors during training.}
    \label{ablation:resnet50:latency}
    \vspace{-0.5em}
  \end{table}

  \begin{table}
    \footnotesize
    \centering
    \begin{tabular}{crrc} \hline
      Network & Params & FLOPs & Error (\%) \\ \hline
      \multirow{4}{*}{\shortstack{SP-ResNet-50 \\(w/o zpm pruning)}} & 25.6M & 4.1G & 24.7 \\
                                                                           & 14.6M & 2.3G & 25.8 \\
                                                                           &  6.7M & 1.1G & 29.8 \\
                                                                           &  1.8M & 286M & 39.6 \\ \hline
      \multirow{4}{*}{\shortstack{SP-ResNet-50 \\(w/ zpm pruning)}}  & 25.6M & 4.1G & 24.2 \\
                                                                           & 17.6M & 2.3G & 24.8 \\
                                                                           &  9.4M & 1.1G & 27.4 \\
                                                                           &  3.8M & 286M & 38.2 \\ \hline
      \multirow{4}{*}{\shortstack{SP-ResNet-50 \\(+ multi-base pruning)}}  & 25.6M & 4.1G & \textbf{23.8} \\
                                                                           & 17.6M & 2.3G & \textbf{24.6} \\
                                                                           & 10.3M & 1.1G & \textbf{26.7} \\
                                                                           &  4.0M & 286M & \textbf{35.0} \\ \hline
    \end{tabular}
    \vspace{-0.5em}
    \caption{Performance comparison between with or without zpm pruning (and +multi-base pruning) with top-1 test errors of ResNet-50 on ImageNet. Prefix SP indicates SP-Net. Classification errors are reported from best test errors during training.}
    \label{ablation:resnet50:zpm}
  \end{table}

  \section{Conclusion}
  In this work, we have proposed multi-base pruning, scs, and zpm pruning to boost the performance of S-Net. With our methods, we realized high performance on par with various NAS models while keeping its flexibility to change computing capacity dynamically like S-Net.

  \clearpage
  {\small
  \bibliographystyle{ieee_fullname}
  \bibliography{egbib}
  }

  \appendix
  \appendixpage
  \section{Implementation Details}
  \label{appnd:implement}
  In this section, we show the implementation details of adaptive computation of S-Net \cite{yu:slimmable} and SP-Net. We also show the implementation details of zpm pruning, and explain about network slimming \cite{liu:slimming}, which is used as a pruning method of SP-Net in our experiments.

  \subsection{S-Net}
  \label{appnd:snet}
  S-Net defines multiple sub-networks as its adaptive computational routes, and architectures of these sub-networks are composed by selecting predefined proportions of channels to use in each layer dynamically. As implementation, S-Net uses slice operations \([:]\) used in python-like style. For instance, considering a convolutional layer \(l\, (l = 1, \ldots, L)\) of \(H \times W\) filter size with \(N_l\) filters, weight parameters of the filter can be denoted as \(\mathbf{W} \in \mathbb{R}^{N_l \times N_{l-1} \times H \times W}\) where \(N_0\) denotes the number of channels of input data. If \(p_l \in (0, 1]\) denotes what proportions of channels to use in layer \(l\), and \(\mathcal{X}^l\) denotes inputs to layer \(l\), outputs of layer \(l\) of S-Net can be denoted as
  \begin{equation}
    \mathcal{Y}^l = \mathbf{W}[:\!p_lN_l,\; :\!p_{l-1}N_{l-1},\; :,\; :]\ast\mathcal{X}^l
  \end{equation}
  S-Net predefines width (proportion of channels) list
  \begin{equation}
    W = [w_1, \ldots, w_k]
  \end{equation}
  Then S-Net uses each width of width list as width multiplier \cite{howard:mobilenets,sandler:mobilenetv2}.
  \begin{equation}
    p_0 = 1,\; p_1 = p_2 = \cdots = p_L = w_i \in W
  \end{equation}
  Therefore, S-Net uses specified proportion (\(w_i \in W\)) of channels in each layer uniformly and dynamically by selecting one of width settings from \(W\) according to the device states.

  \subsection{SP-Net}
  \label{appnd:spnet}
  With multi-base pruning, we first define width list the same as S-Net
  \begin{equation}
    W = [w_1, \ldots, w_k]
  \end{equation}
  Then we train individual base networks with specified proportion (\(w_i \in W \mysetminus \{w_1\}\)) of channels to use in each layer. For each of the trained base networks \(M_{w_i}\), we prune it and obtain pruned architectures \(\mathcal{A}_{w_{i-1}}\) of the same FLOPs as \(M_{w_{i-1}}\). If \(N^l_{w_{j}} (w_j \in W \mysetminus \{w_k\})\) denotes the number of remained filters in layer \(l\) after pruning \(M_{w_{j+1}}\), the pruned architectures are denoted as
  \begin{equation}
    \mathcal{A}_{w_{j}} = \{N_0, N^1_{w_j}, N^2_{w_j}, \ldots, N^L_{w_j}\}\;\; (w_j \in W \mysetminus \{w_k\})
  \end{equation}
  Then we embed \(\{\mathcal{A}_{w_1}, \ldots, \mathcal{A}_{w_{k-1}}\}\) into \(M_{w_k}\) as its sub-networks. After sorting filters and input channels in each layer of \(M_{w_k}\) following scs, outputs of layer \(l\) of SP-Net can be denoted as
  \begin{equation}
    \mathcal{Y}^l = \mathbf{W}[:\!N^l_{w_j},\; :\!N^{l-1}_{w_j},\; :,\; :]\ast\mathcal{X}^l,\;\; w_j \in W
  \end{equation}
  where \(N^l_{w_k} = N_l,\; N^0_{w_j} = N_0\). Therefore, SP-Net chooses one of embedded pruned architectures adaptively and calculates outputs by selecting one of width settings from \(W\) according to the device states.

  \subsection{zpm pruning}
  \label{appnd:zpm}
  As described in Sec. 3.4, this method keeps the pruning pattern on both paths of residual structures while filling missing channels with zeros, then adds up outputs from these paths. As implementation, we prepare zero tensors of the same size as before-pruned channels of these both paths, then use the tensors as base of add operations of outputs from these paths (do add operations on this zero tensors).

  However, this method requires additional implementation when we use it with slimmable channel sorting (scs), because scs sorts filters in each layer according to pruning precedence. When we use scs on residual structures, it sorts channels differently between shortcut and non-shortcut path since these paths are pruned differently. Thus add operations occur between different channel index with zpm pruning (\eg between channel index 1 of shortcut path and channel index 5 of non-shortcut path). Therefore, we introduced two additional tensors to solve this issue: (1) masks which stores information that which channels were selected by pruning before sorting, and (2) indices which restores original orders of channels before sorting. On calculating added outputs from confluence of residual structures, we first prepare zero tensors same as the case without scs, and restores original orders of channels with the indices, then do add operations of each (shortcut and non-shortcut) path on the zero tensors where masks have non-zero bits.

  \subsection{network slimming}
  \label{appnd:pruning}
  Network slimming \cite{liu:slimming} is one of the channel pruning methods. It prunes channels corresponding to smaller absolute values of weight parameters in Batch Normalization \cite{ioffe:bn} layer. This method introduces additional process of updating weights so that it minimizes following objective function.

  \begin{equation}
    L = \sum_{(x, y)} l\left(f(x, \mathbf{W}),\; y\right) + \lambda\sum_{\gamma \in \Gamma}g(\gamma),
  \end{equation}

  where \((x, y)\) denotes a pair of data and labels, \(\mathbf{W}\) weight parameters of network \(f\), \(l\) loss function, \(\Gamma\) set of weight parameters of all the Batch Normalization layers, \(\gamma\) weight parameters of a Batch Normalization layer, \(g(\cdot)\) sparsity-inducing function, and \(\lambda\) balancing hyperparameter of two terms. First term represents original training loss of \(f\). Second term represents regularization term on weight parameters of Batch Normalization layer. In order to promote sparsity of weight parameters, Liu \etal~\cite{liu:slimming} used \(g(s) = |s|\), and subgradient descent is used for optimizing non-smooth \(L_1\) term.

  \section{Ablation Study}

  \subsection{Comparison of pruning algorithms}
  \label{appnd:pruningcomparison}
  We can use any channel pruning algorithms for SP-Net, since any channel-pruned structures can be embedded into \(\times\)1.0 base network. Through whole experiments in this paper, we used \textit{network slimming} \cite{liu:slimming} as a pruning method of SP-Net, because it's easy to implement and effective. We also show the results using \(L_1\)-norm based pruning as a pruning method of SP-Net in \cref{ablation:resnet50:pruningofspnet}.

  \begin{table}[H]
    \footnotesize
    \centering
    \begin{tabular}{>{\centering\arraybackslash}m{5.8em} >{\centering\arraybackslash}m{7em} >{\hspace*{0pt}\hfill}m{3.6em} >{\hspace*{0pt}\hfill}m{3.4em} >{\centering\arraybackslash}m{2.3em}} \Xhline{2\arrayrulewidth}
      Model & Pruning Method & Params & FLOPs & \vspace{0.2em}\shortstack{Error \\ (\%)} \\ \Xhline{2\arrayrulewidth}
      \multirow{12}{*}{ResNet-50}  & \multirow{3}{*}{\shortstack{SP (w/ network \\ slimming \cite{liu:slimming})}}
                                     & 25.6M & 4.1G & \textbf{23.3} \\
                                   & & 16.0M & 2.0G & \textbf{24.3} \\
                                   & & 10.2M & 1.0G & \textbf{26.0} \\ \cline{2-5}
                                   & \multirow{3}{*}{\shortstack{network \\ slimming \cite{liu:slimming}}}
                                     & 25.6M & 4.1G & 23.6 \\
                                   & & 13.6M & 2.3G & 25.5 \\
                                   & &  6.8M & 1.1G & 31.8 \\ \cline{2-5}
                                   & \multirow{3}{*}{\shortstack{SP (w/ \(L_1\)-norm \\ based pruning \cite{li:convnets})}}
                                     & 25.6M & 4.1G & 24.2 \\
                                   & & 20.9M & 2.3G & 25.1 \\
                                   & & 10.0M & 1.1G & 28.3 \\ \cline{2-5}
                                   & \multirow{3}{*}{\shortstack{\(L_1\)-norm based \\ pruning \cite{li:convnets}}}
                                     & 25.6M & 4.1G & 23.4 \\
                                   & & 21.2M & 2.3G & 25.2 \\
                                   & & 10.6M & 1.1G & 31.5 \\ \Xhline{2\arrayrulewidth}
    \end{tabular}
    \vspace{-0.5em}
    \caption{Performance comparison between network slimming \cite{liu:slimming} and \(L_1\)-norm based pruning \cite{li:convnets} as pruning methods of SP-Net with top-1 test errors on ImageNet. Results of network slimming and \(L_1\)-norm based pruning are reproduced and reported from best test errors during training.}
    \label{ablation:resnet50:pruningofspnet}
  \end{table}

  From comparison between network slimming and \(L_1\)-norm based pruning, we see that \(L_1\)-norm based pruning itself outperforms network slimming, but with our methods, network slimming outperforms \(L_1\)-norm based pruning significantly. Thus it is not trivial whether one pruning algorithm with SP-Net surpasses performance of other methods with SP-Net even if it outperforms them when used alone.

  \subsection{Comparison of hardware latencies}
  In this paper, we only showed the comparison of inference latencies of ResNet-50. In this section, we also show latency comparison of other models between S-Net and SP-Net.

  \begin{table}[H]
    \footnotesize
    \centering
    \begin{tabular}{>{\centering\arraybackslash}m{5.8em} >{\centering\arraybackslash}m{3em} >{\hspace*{0pt}\hfill}m{2.8em} >{\hspace*{0pt}\hfill}m{2.4em} >{\centering\arraybackslash}m{2.3em} >{\centering\arraybackslash}m{3.3em}} \Xhline{2\arrayrulewidth}
      Model & Method & Params & FLOPs & Error (\%) & Latency \\ \Xhline{2\arrayrulewidth}
      \multirow{6}{*}{VGGNet}      & \multirow{3}{*}{S-Net}
                                     & 20.6M & 19.6G & 27.0 & 103.5ms \\
                                   & &  5.3M &  4.9G & 31.5 & 37.0ms \\
                                   & &  1.4M &  1.2G & 41.4 & 15.8ms \\ \cline{2-6}
                                   & \multirow{3}{*}{\shortstack{SP-Net \\ (ours)}}
                                     & 20.5M & 19.6G & \textbf{26.1} & 103.5ms \\
                                   & &  6.6M &  4.9G & \textbf{29.3} & 40.9ms \\
                                   & &  1.8M &  1.2G & \textbf{37.0} & 18.0ms \\ \Xhline{2\arrayrulewidth}
      \multirow{6}{*}{MobileNetV1} & \multirow{3}{*}{S-Net}
                                     &  4.3M &  584M & 28.6 & 16.7ms \\
                                   & &  2.6M &  337M & 30.5 & 11.5ms \\
                                   & &  1.3M &  157M & 34.7 &  7.3ms \\ \cline{2-6}
                                   & \multirow{3}{*}{\shortstack{SP-Net \\ (ours)}}
                                     &  4.9M &  569M & \textbf{27.3} & 17.6ms \\
                                   & &  3.2M &  325M & \textbf{29.0} & 11.9ms \\
                                   & &  1.8M &  150M & \textbf{32.0} &  7.6ms \\ \Xhline{2\arrayrulewidth}
      \multirow{6}{*}{MobileNetV2} & \multirow{3}{*}{S-Net}
                                     &  5.4M &  509M & 26.6 & 25.3ms \\
                                   & &  3.5M &  301M & 28.5 & 18.2ms \\
                                   & &  2.6M &  209M & 30.8 & 15.8ms \\ \cline{2-6}
                                   & \multirow{3}{*}{\shortstack{SP-Net \\ (ours)}}
                                     &  5.4M &  509M & \textbf{25.2} & 25.0ms \\
                                   & &  4.0M &  305M & \textbf{26.5} & 17.8ms \\
                                   & &  2.9M &  207M & \textbf{27.7} & 14.0ms \\ \Xhline{2\arrayrulewidth}
    \end{tabular}
    \vspace{-0.5em}
    \caption{Hardware latencies comparison of inference by various models on NVIDIA V100 GPU with batch size 64 between S-Net and SP-Net. Top-1 test errors of classification on ImageNet are reported from best test errors during training.}
    \label{ablation:latency}
  \end{table}

\end{document}